\documentclass{article}     

\usepackage{arxiv}
\usepackage[utf8]{inputenc} 
\usepackage[T1]{fontenc}    
\usepackage{hyperref}       
\usepackage{url}            
\usepackage{booktabs}       
\usepackage{amsfonts}       
\usepackage{nicefrac}       
\usepackage{microtype}      
\usepackage{lipsum}
\usepackage{graphicx}
\usepackage{caption}
\usepackage{subcaption}
\usepackage{multirow}

\usepackage{amsthm}
\usepackage{bm}
\usepackage{amsmath,amssymb,amsfonts}
\usepackage{algorithm}
\usepackage{algpseudocode}
\title{Optimizing Robotic Manipulation with Decision-RWKV: \\A Recurrent Sequence Modeling Approach for Lifelong Learning}
\author{
  Yujian Dong\\
  Department of Mechanical and Energy Engineering\\
  Southern University of Science and Technology\\
  Shenzhen, China 518055\\
  \And
  Tianyu Wu\\
  Department of Mechanical and Energy Engineering\\
  Southern University of Science and Technology\\
  Shenzhen, China 518055\\
  \And
  Chaoyang Song\thanks{Corresponding Author.}\\
  Design \& Learning Research Group\\
  Southern University of Science and Technology\\
  Shenzhen, China 518055\\
  \texttt{songcy@ieee.org}\\
}
\begin{document}
\maketitle
\begin{abstract}

    Models based on the Transformer architecture have seen widespread application across fields such as natural language processing, computer vision, and robotics, with large language models like ChatGPT revolutionizing machine understanding of human language and demonstrating impressive memory and reproduction capabilities. Traditional machine learning algorithms struggle with catastrophic forgetting, which is detrimental to the diverse and generalized abilities required for robotic deployment. This paper investigates the Receptance Weighted Key Value (RWKV) framework, known for its advanced capabilities in efficient and effective sequence modeling, and its integration with the decision transformer and experience replay architectures. It focuses on potential performance enhancements in sequence decision-making and lifelong robotic learning tasks. We introduce the Decision-RWKV (DRWKV) model and conduct extensive experiments using the D4RL database within the OpenAI Gym environment and on the D’Claw platform to assess the DRWKV model's performance in single-task tests and lifelong learning scenarios, showcasing its ability to handle multiple subtasks efficiently. The code for all algorithms, training, and image rendering in this study is open-sourced at \href{https://github.com/ancorasir/DecisionRWKV}{https://github.com/ancorasir/DecisionRWKV}. 
    
\end{abstract}
\keywords{
    Lifelong Learning \and Robot Learning \and Robotic Manipulation \and Sequence Modeling
}   
\section{Introduction}
\label{sec:Intro}
    
    Artificial intelligence has witnessed a paradigm shift in recent years with the advent of large language models (LLMs), particularly those based on the Transformer architecture~\cite{vaswani2017attention}. These models have revolutionized how machines understand and generate human language, significantly advancing natural language processing (NLP) and extending their influence to computer vision~\cite{dosovitskiy2020image, liu2021swin}. The versatility of Transformer-based models is further underscored by their successful deployment as autonomous agents and decision-makers in robotics~\cite{shridhar2023perceiver}, where they have been instrumental in enabling machines to perform complex tasks with a degree of autonomy previously unattainable.
        
    With the advancement of LLM, an intriguing development has been integrating these models with robotic motion, utilizing LLMs as planners to decompose high-level commands into executable instruction sequences for robotic motion control~\cite{parakh2023lifelong}. This approach has demonstrated promising results, showcasing the versatility of LLMs beyond traditional natural language processing tasks. Similarly, the application of Transformers, which has achieved remarkable success in NLP tasks, has seen a comparable surge in reinforcement learning (RL)~\cite{li2023survey}. However, adopting Transformers within Reinforcement Learning (RL) introduces unique design choices and challenges stemming from RL's intrinsic attributes. As discussed in the context of the Decision Transformer (DT)~\cite{chen2021decision}, these challenges necessitate meticulous consideration of the adjustments to the Transformer model to ensure its efficacy within RL settings.
    
    Traditional machine learning algorithms have long grappled with the issue of catastrophic forgetting~\cite{kirkpatrick2017overcoming}, a phenomenon where the model's performance on previously learned tasks rapidly deteriorates as it acquires new knowledge~\cite{mccloskey1989catastrophic}. This poses a severe obstacle to realizing lifelong learning systems, which are expected to learn continuously from a data stream and accumulate knowledge over time~\cite{parisi2019continual}.In light of the remarkable memory retention and generalization aptitudes exhibited by Transformer-based models in many domains, there is a burgeoning interest in creating a model architecture capable of facilitating multi-task continual learning without imposing substantial computational burdens. Such an advancement would substantially propel the progression of lifelong robot learning.
    
    Recently, the Receptance Weighted Key Value (RWKV)~\cite{peng-etal-2023-rwkv, peng2024eagle} framework has been introduced as a highly efficient and practical framework for sequence modeling, which combines the computational complexity advantages of RNNs with the parallel training capabilities of Transformers. The framework constructs model blocks by integrating Time mix and Channel mix techniques, and its efficacy has been successfully validated on large-scale models.
    
\begin{figure}[htbp]
        \centering
        \includegraphics[width=0.9\linewidth]{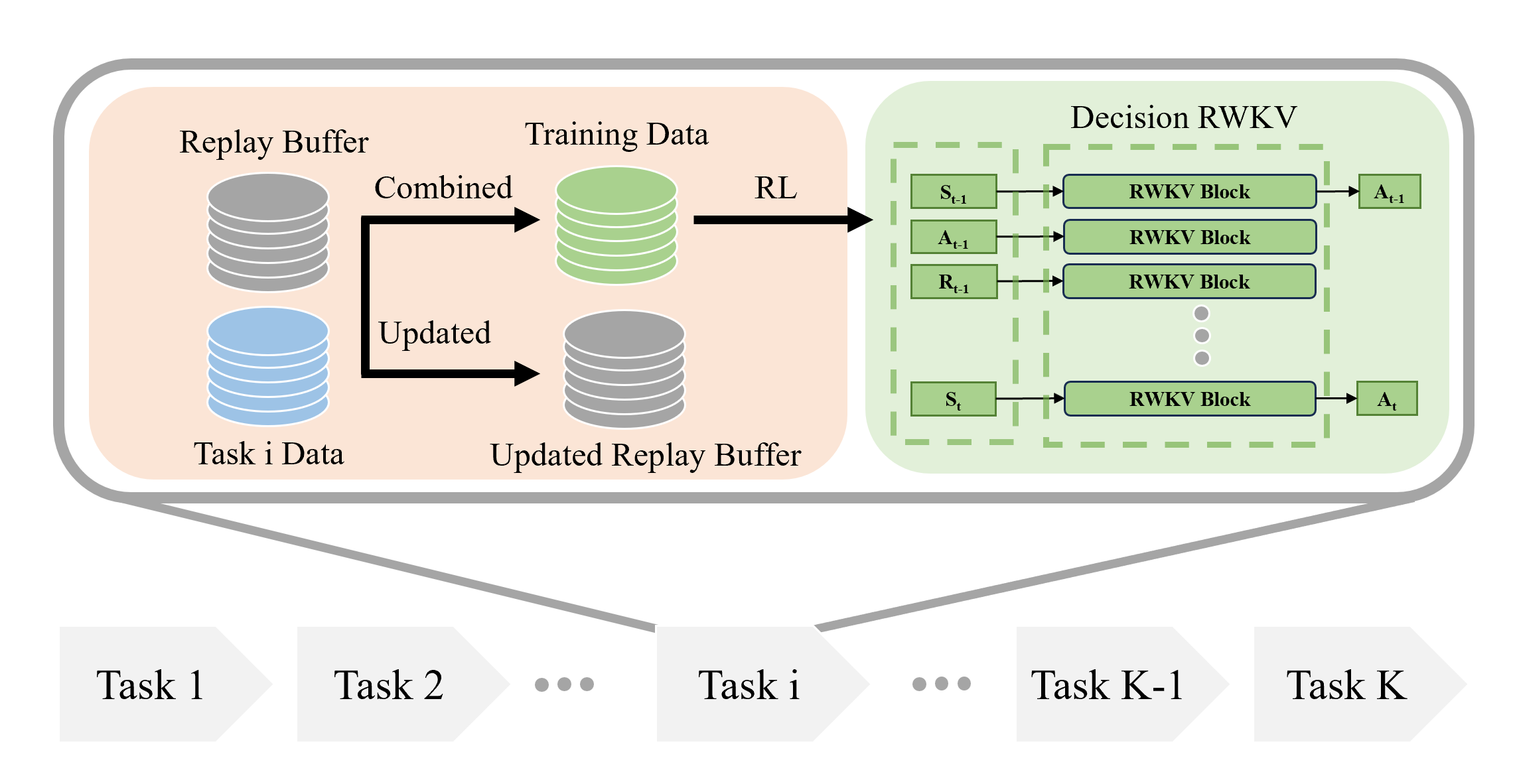}
        \caption{An overview of our algorithm architecture. For a single task i, we employ RWKV blocks for sequential decision-making learning. During continual learning across multiple tasks, we maintain the model's memory capabilities across various tasks by incorporating a replay buffer to utilize the experience replay method.}
        \label{fig:Intro_PaperOverview}
    \end{figure}

    This paper introduces the Decision-RWKV (DRWKV) model as a sequence decision-making model, as shown in Fig.~\ref{fig:Intro_PaperOverview}. To enhance memory capabilities during multi-task or lifelong robot learning processes, we incorporate an experience replay module. We conducted individual task tests using tasks from the D4RL dataset. Additionally, we constructed an offline dataset based on the D'Claw platform, which includes ten similar valve-turning tasks, to evaluate the decision model's performance in the context of lifelong robot learning. 
    
    Our experimental evaluation focuses on two aspects. Firstly, we investigate the model's ability to continuously learn across ten sub-tasks in a lifelong learning scenario, where each task is encountered only once and not revisited after completion. We assess whether there is catastrophic forgetting of previous tasks after learning new ones, testing the model's lifelong learning capabilities. Secondly, stability and low latency are prerequisites for robotic end-effectors towards successful human-robot interaction. We emphasize testing the computational resources and latency incurred while performing task tests using the dataset. In summary, our paper provides the following three contributions:
    \begin{enumerate}
        \item Constructed an offline dataset for lifelong robot learning evaluation based on the D'Claw platform.
        \item Presented the Decision RWKV as a novel architecture that enhances lifelong learning capabilities in robots by integrating sequence modeling with robotic reinforcement learning.
        \item Assessed the performance of Decision RWKV in complex lifelong learning tasks, emphasizing the practical trade-offs and challenges when deploying such architectures in real scenarios.
    \end{enumerate}

\section{Related Work}

\subsection{Related Work on Foundation Models}

    Foundation models (FMs), or large models pre-trained on massive data and adapted for downstream tasks, have emerged as a practical paradigm in modern machine learning. Represented by models like ChatGPT~\cite{kocon2023chatgpt, qin2023chatgpt}, these foundational models have marked a significant advancement in artificial intelligence, particularly impacting natural language processing and generation. Built upon the architecture of Transformer~\cite{vaswani2017attention} and its core attention layer~\cite{bahdanau2014neural}, this development has fundamentally transformed the capabilities of artificial intelligence.

    The efficacy of self-attention is attributed to its ability to densely route information within a context window, enabling it to model complex data. However, this property brings fundamental drawbacks: an inability to model anything outside a finite window and quadratic scaling for sequence length. A substantial amount of research has emerged to explore more efficient variants of attention to overcome these drawbacks, such as mamba~\cite{gu2023mamba}, retnet~\cite{sun2023retentive}, RWKV~\cite{peng-etal-2023-rwkv, peng2024eagle}, aiming to reduce the computational resource requirements needed. This limitation is particularly crucial for using these foundational models in end devices, especially in robotics, where computational resources are constrained.

\subsection{Related Work on RWKV}
    
    The RWKV framework has emerged as a powerful alternative to conventional models in NLP and computer vision~\cite{duan2024vision}. The RWKV model is precisely engineered to address the challenges associated with processing long sequences, a crucial aspect of robotic tasks that require sustained temporal dependencies. Unlike traditional Transformers, which suffer from quadratic complexity as sequence length increases, the RWKV model is characterized by its linear complexity. This key feature significantly enhances the model's scalability and adaptability for real-time applications where processing efficiency is paramount.

    Furthermore, the RWKV framework demonstrates exceptional memory retention capabilities, a vital attribute for lifelong learning applications. By efficiently building upon previously acquired knowledge without falling into the common pitfalls of memory degradation over time, RWKV stands out as a robust solution for continuous learning scenarios. Additionally, the architecture of RWKV is inherently compatible with the reinforcement learning paradigm. This compatibility facilitates the seamless integration of decision-making processes within the sequence modeling tasks, enhancing the model's utility in adaptive and dynamic environments.

    Another significant advantage of the RWKV model over traditional Recurrent Neural Networks (RNNs) is its ability to overcome the limitations typically associated with RNNs, such as difficulties in parallelization and the notorious issues of vanishing or exploding gradients. These enhancements not only improve computational efficiency but also ensure more stable and reliable performance across a range of applications. These attributes make the RWKV framework an attractive choice for researchers and practitioners seeking efficient and effective solutions in sequence modeling.

\subsection{Related Work on Sequential Decision Model}

    Sequential decision-making problems are central to Reinforcement Learning (RL). Researchers have been continuously exploring how to solve these problems more effectively in recent years. Traditional RL approaches often rely on complex dynamic programming algorithms and value function approximation. However, recent studies suggest that reconceptualizing RL problems as sequence modeling tasks can be a more intuitive and flexible approach.

    Offline learning in reinforcement learning refers to training agents using datasets collected beforehand without requiring real-time interaction with the environment. This learning mode is beneficial for scenarios where online data collection is costly or dangerous. In contrast, online learning involves continuous interaction with the environment during the learning process, allowing the agent to adapt to new situations and data distributions in real time.

    The original DT~\cite{chen2021decision} exemplifies this novel perspective. It transforms the RL problem into a sequence prediction task by learning a policy capable of predicting the action given a state and cumulative reward. The advantage of this approach is its natural compatibility with established sequence modeling techniques, such as the Transformer architecture, to handle RL tasks. The Online DT~\cite{zheng2022online} combines an offline pre-trained DT model with an online fine-tuning process. This method aims to address the distribution shift problem common in offline RL, where the performance of pre-trained models often degrades in actual online environments due to changes in data distribution. Online fine-tuning allows the model to adapt to new data distributions, improving its practical application performance. In multi-agent settings, similar online DT methods have been proposed~\cite{lee2022multi}. In such setups, models must learn to make decisions while interacting with other agents, adding complexity to the problem. The introduction of multi-agent online DT methods underscores the potential of sequential decision-making approaches in broader application scenarios.

    Furthermore, the transfer learning capability of the DT architecture has been explored~\cite{reid2022can}. DT has demonstrated its cross-domain and cross-modality generalization abilities by pre-training on diverse domains and modalities, such as Wikipedia articles, and then fine-tuning on specific offline RL tasks. This capability has been validated on a large scale in Gato~\cite{reed2022generalist}, where DT exhibited impressive performance across various tasks and modalities. Given the substantial computational demands of DT, the work on DLSTM~\cite{siebenborn2022crucial} has also begun to explore modifications to the Transformer architecture within DT regarding inference time and real-time response performance.

    These research findings indicate that the decision model is broadly applicable and has potential in both single-agent and multi-agent environments and both online and offline learning settings. They reveal a universal structure within sequential decision-making problems, enabling cross-domain and multi-modal transfer learning. This provides a fresh perspective for tackling complex RL challenges and directs future research toward exploring and exploiting this universal structure to develop more robust and generalizable sequential decision-making algorithms.

\subsection{Related Work on Lifelong Robot Learning}

    Lifelong learning is essential for agents to perpetually acquire and adapt to new knowledge across diverse tasks and environments, effectively mitigating the issue of catastrophic forgetting. This process necessitates the seamless transfer and accumulation of knowledge without compromising previously acquired information. Contemporary algorithms in lifelong learning can be systematically classified into four principal categories.

    \textbf{Regularization-Based Methods:} These approaches incorporate additional regularization terms into the loss function to safeguard vital information about antecedent tasks. A notable example, the Elastic Weight Consolidation (EWC) method~\cite{kirkpatrick2017overcoming}, leverages the Fisher information matrix to ascertain and preserve the stability of critical network parameters, thereby minimizing the loss of pre-existing knowledge during the acquisition of new tasks. A significant advantage of this methodology is maintaining a constant model size irrespective of the increment in task numbers.

    \textbf{Memory-Based Methods:} By employing a replay buffer, these methods archive and subsequently replay samples from previous tasks to aid in learning new tasks. Experience Replay (ER)~\cite{schaul2015prioritized} epitomizes this strategy by allowing agents to reutilize past experiences, enhancing learning efficiency by disrupting data time correlations.

    \textbf{Gradient-Constraint Methods:} These methods are dedicated to attenuating the detrimental impacts of learning new tasks on preceding tasks via gradient adjustments. Gradient Episodic Memory (GEM)~\cite{lopez2017gradient} and its computationally optimized counterpart, A-GEM~\cite{chaudhry2018efficient}, project gradients into spaces that avert performance deterioration on older tasks. Orthogonal Gradient Descent (OGD)~\cite{farajtabar2020orthogonal} minimizes interference by orthogonally updating parameters relative to the gradients of prior tasks.

    \textbf{Expansion-Based Methods:} This category introduces new submodules for each incoming task, culminating in a progressively expanding model size. Additive Parameter Decomposition (APD)~\cite{yoon2019scalable} integrates task-specific subnetworks while conserving a shared parameter space, minimizing modifications to shared parameters by introducing new tasks.

    However, applying these algorithms in the real world, especially within robotic agents, encounters various challenges. Lifelong robot learning~\cite{thrun1995lifelong} aims to address this issue. It has been manifested in applications such as mobile robot navigation~\cite{liu2021lifelong} and robotic manipulation tasks~\cite{xie2022lifelong}, demonstrating the advantages of lifelong robot learning in real robotic applications. The current research direction for lifelong robot learning still focuses on designing better neural architectures, improving algorithms to enhance positive transfer capabilities, and utilizing pre-training to boost lifelong learning performance~\cite{liu2024libero}.
    
\section{Preliminaries}
\label{sec:Prelim}

\subsection{Markov Decision Processes in Robotic Learning}

    The problem of learning in robotics can be modeled as a finite-horizon Markov Decision Process (MDP): $\mathcal{M} = (S, A, \mathcal{T}, H, \mu_0, R)$. In this model, $S$ and $A$ denote the spaces of states and actions available to the robot. $\mu_0$ represents the distribution of initial states, $R : S \times A \rightarrow \mathbb{R}$ is the function that assigns rewards, and $\mathcal{T} : S \times A \rightarrow S$ describes the state transition dynamics. In our approach, we consider a scenario where rewards are sparse and thus substitute $R$ with a binary goal indicator $g: S \rightarrow \{0, 1\}$. The goal for the robot is to devise a policy $\pi$ that is optimal in terms of maximizing the expected cumulative reward:
    \begin{equation}
        \max_\pi J(\pi) = \mathbb{E}_{s_t,a_t \sim \pi,\mu_0} \left[ \prod_{t=1}^{H} g(s_t) \right]
    \end{equation}
    
\subsection{Lifelong Learning in Robotics}

    In the context of lifelong learning for robotics, a robot sequentially masters a series of $K$ tasks, $\{T_1, \ldots, T_K\}$, through a unified policy $\pi$. This policy is designed to be task-aware, i.e., $\pi(\cdot \mid s; T)$. Each task $T_k$ is characterized by its unique initial state distribution $\mu^k_0$ and a specific goal indicator $g^k$. It is assumed that the sets of states $S$, actions $A$, transitions $\mathcal{T}$, and the horizon $H$ remain consistent across all tasks. The objective for the robot, up to and including the $k$-th task $T_k$, is to refine its policy $\pi$ to maximize the average performance across all learned tasks\cite{liu2024libero}:
    \begin{equation}
        \label{equ:lifelongRobotLearning}
        \max_{\pi} J_{\text{LRL}}(\pi) = \frac{1}{k} \sum_{p=1}^{k} \left( \mathbb{E}_{s^p_t, a^p_t \sim \pi(\cdot;T_p), \mu^p_0} \left[ \prod_{t=1}^{H} g^p(s^p_t) \right] \right). 
    \end{equation}
    
\section{Methods}
\label{sec:Methods}

    As illustrated in Fig.~\ref{fig:Intro_PaperOverview}, we employ a Decision Blocks approach for individual tasks inspired by DT to model and predict serialized task information to complete control tasks. In the context of lifelong robot learning tasks, it is necessary to consider the continuous learning process across different tasks. We initialize a fixed-size replay buffer to sample the model's learned data from subsequent tasks. When the model learns from the \textit{i}-th task, the replay buffer is combined with the current task's offline dataset to form the training dataset. After the task learning is completed, the replay buffer is merged with the current task dataset to prepare the data for the next task. Through this approach, while maintaining the model configuration, we effectively preserve memory across different tasks to prevent catastrophic forgetting.

\subsection{Lifelong Robot Learning as a Sequential Model Problem}

    Learning can be effectively facilitated through a sequence prediction model that encapsulates the robot motion process in a serialized format. As depicted in Eq.~\eqref{eq:tra}, the key variables—Return, State, and Action—are structured chronologically.

    Eq.~\eqref{eq:tra} delineates the sequence prediction model, which is pivotal to our strategy for lifelong robot learning. This model processes the robot's motion as a temporal sequence, integrating Return, State, and Action. Adopting this sequence-based methodology allows the robot to assimilate knowledge from a continuum of experiences. This capability is crucial for the robot's adaptation to new tasks while preserving insights from previous undertakings, thus exemplifying the principles of lifelong learning.

    \begin{align}
        \label{eq:tra}
        \tau = \left( \hat{R_{1}}, s_{1}, a_{1},\hat{R_{2}}, s_{2}, a_{2}, \ldots, \hat{R_{T}}, s_{T}, a_{T} \right)
    \end{align}
    
\subsection{Decision RWKV}
    
    RWKV-4~\cite{peng-etal-2023-rwkv} represents an innovative model architecture that merges Transformers' efficient, parallelizable training capabilities with the compelling inference advantages of Recurrent Neural Networks. The model employs stacked residual blocks, each consisting of time-mixing and channel-mixing sub-blocks, forming a recurrent structure to leverage past information. Time and channel mixing are orchestrated through specific weight decay and update mechanisms. Concurrently, RWKV-V4 utilizes a variant of the linear attention mechanism, replacing the traditional dot-product attention mechanism with a more efficient channel-directed attention approach. 

    RWKV-5 and RWKV-6~\cite{peng2024eagle} are advanced iterations of the RWKV model architecture; RWKV-5 introduces multi-head matrix-valued states, enhancing the model's expressive capacity. RWKV-6 incorporates new data-dependent functions for time and token mixing modules, increasing the model's flexibility and expressiveness. The WKV attention submodule in RWKV-5 employs learnable per-channel decay rates. At the same time, RWKV-6 further introduces the Low-Rank Adaptation (LoRA) mechanism, allowing the model to dynamically generate token shift amounts and decay rates in a data-driven manner. Both new models utilize SiLU gating instead of the sigmoid activation function for output gating, boosting the model's performance. In the LLM (Large Language Models) domain, the significant performance enhancements brought about by the RWKV model mechanism updates raise an intriguing question: whether this relationship could also correspond to performance improvements in decision models.

\begin{figure}[htbp]
        \centering
        \includegraphics[width=0.9\linewidth]{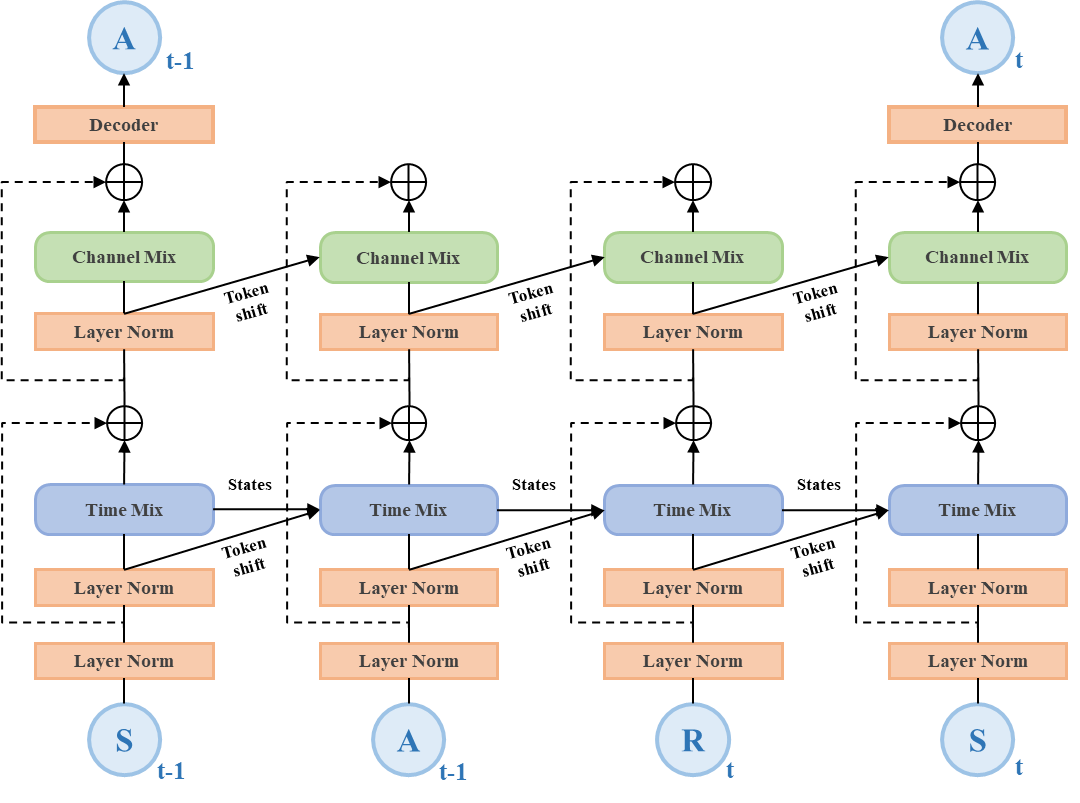}
        \caption{The structure of the DRWKV model, where A represents action, referring to the robot's motion information; S represents state, indicating the current state information; R stands for return, which is the reward information obtained. We introduce DRWKV by utilizing the RWKV block, which incorporates Time-mix and Channel-mix, as a token-mixing module instead of the self-attention module in DT.}
        \label{fig:Method_DRWKV}
    \end{figure}

    As illustrated in Fig.~\ref{fig:Method_DRWKV}, we introduce Decision RWKV (DRWKV) by utilizing the RWKV block, which incorporates Time-mix and channel-mix, as a token-mixing module instead of the self-attention module of DT.

\subsection{Incorporating Experience Replay into the Decision Model for Lifelong Robot Learning}

    Incorporating Experience Replay into the Decision Model for Lifelong Robot Learning enables robotic systems to achieve the capability for lifelong learning. This is achieved by refining the integration of the Decision RWKV model with an Experience Replay mechanism. This strategic combination supports uninterrupted learning across diverse tasks and settings. It allows robots to progressively enhance their decision-making skills by leveraging accumulated experiences.

    Experience Replay plays a pivotal role in this process. It is a technique where an agent stores its experiences represented as state-action-reward-next state (SARS) tuples in a replay buffer. These experiences are randomly sampled during training, allowing the agent to re-engage with past interactions. This reduces temporal correlations between successive samples and introduces a variety of learning situations. Collectively, these improvements enhance the learning process's stability and effectiveness.

    The Decision RWKV model integrates the RWKV's efficient linear attention mechanism with the Decision Transformer's decision-making prowess. This integration makes the model adept at handling long sequences and complex state spaces. It is highly suitable for robotic control's dynamic and sequential decision-making needs. In the integration process, robots interact with their environment, taking actions and noting the consequent states and rewards. These events are recorded as SARS tuples and stored in the replay buffer.

    A minibatch of experiences is randomly drawn from the replay buffer for training. The Decision RWKV model uses these samples to refine its policy. It predicts optimal actions based on past states and actions while considering expected returns. The model treats SARS tuples as sequences, applying its linear attention mechanism to process these sequences autoregressively. This enables the model to recognize temporal patterns and develop complex behaviors over extended periods.

    As the robot continuously interacts with its environment and gathers new experiences, these are added to the replay buffer. The model is periodically retrained with this updated data, promoting the robot's ability to learn and adapt indefinitely. The integration with Experience Replay allows robots to maximize the educational value of each encounter. Learning from successes and setbacks and using random sampling from the replay buffer disrupts the correlations among experiences. This fosters a more stable and diverse learning process.

    The Decision RWKV model's scalability and proficiency in handling lengthy sequences match the intricacies of lifelong learning in robotics. This combined system enables robots to adapt to environmental changes and progressively acquire new tasks. It exemplifies the essence of lifelong learning. To ensure the model retains a broad spectrum of experiences, each new task's data is incorporated into the replay buffer. The buffer maintains a fixed size of 10,000 tuples, significantly enhancing the model's memory capabilities. It does so by blending task-specific data with the diverse historical data from the replay buffer for training.

    The enhanced integration of the Decision RWKV model with Experience Replay provides a potent framework for supporting lifelong learning in robotic systems. It reinforces the robot's ability to learn from various experiences and guarantees ongoing development and adaptability to novel challenges. Thus furthering the pursuit of intelligent and autonomous robotic agents.

\section{Experiment and Results}
\label{sec:Results}

\subsection{Single-Task Offline Reinforcement Learning}

\begin{figure}[htbp]
        \centering
        \includegraphics[width=1\linewidth]{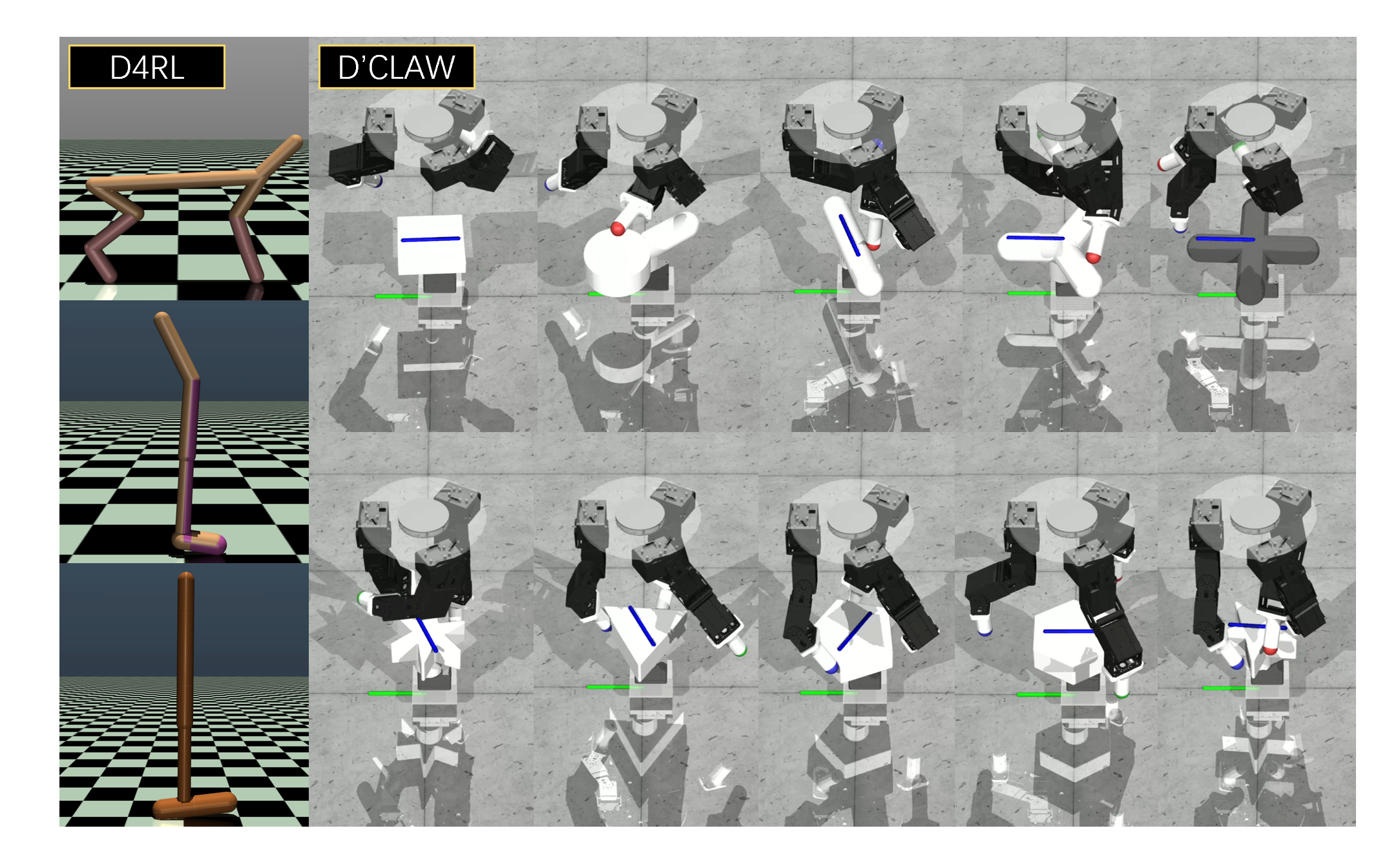}
        \caption{Simulation Experimental Environment Overview. We utilize the D4RL dataset for single-task offline reinforcement learning and the D'Claw dataset for lifelong robot learning tasks.}
        \label{fig:Intro_Dataset}
    \end{figure}

    We utilized the D4RL~\cite{fu2020d4rl} database for continuous control tasks in OpenAI Gym~\cite{brockman2016openai}, which includes several continuous locomotion tasks with dense rewards. Our experiments were conducted in the simulation environments of \textit{HalfCheetah}, \textit{Hopper}, and \textit{Walker} as shown in Fig.~\ref{fig:Intro_Dataset}. 

    For each environment, we tested offline reinforcement learning using datasets of three different quality levels:
    \begin{itemize}
        \item Medium (\textit{m}): one million timesteps generated by a policy that achieves approximately one-third the score of an expert policy.
        \item Medium-Replay (\textit{m-r}): the replay buffer of an agent trained to match the performance of the Medium policy.
        \item Medium-Expert (\textit{m-e}): one million timesteps generated by the Medium policy, supplemented with one million timesteps generated by an expert policy.
    \end{itemize}
    
\subsubsection{Training Progress}

    To guarantee variable consistency across the training of diverse models, we adapted the code from the CORL~\cite{tarasov2024corl} training library and developed training scripts for DT, DRWKV4, DRKWV5, and DRWKV6. The training was conducted on the D4RL dataset, during which we meticulously documented and analyzed the progression of training loss reduction, memory consumption, and computational duration about increasing sequence lengths. This approach was undertaken to furnish robust support for subsequent data analysis endeavors.

\subsubsection{Result Analysis}
    
\begin{figure}[htbp]
        \centering
        \includegraphics[width=0.7\linewidth]{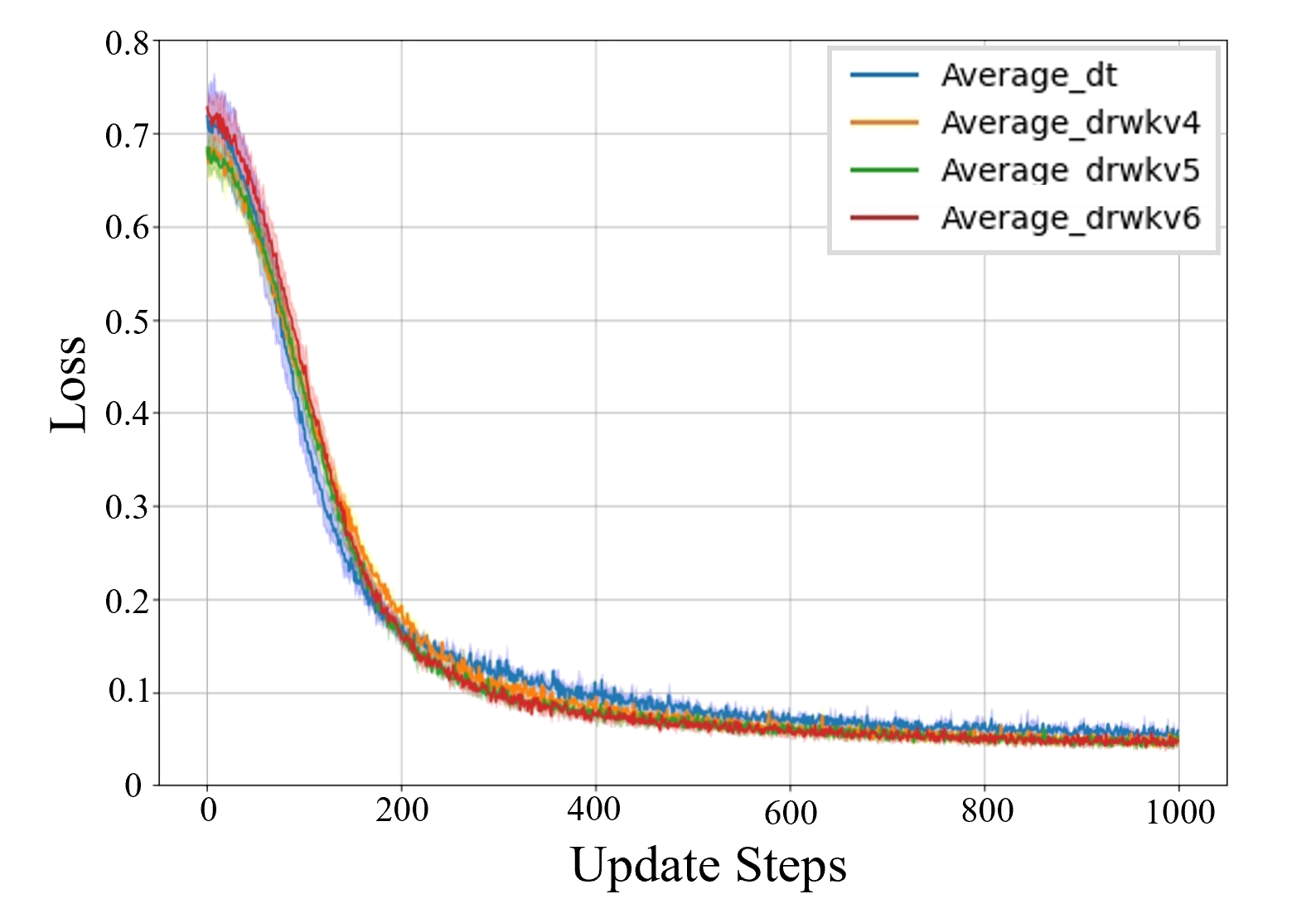}
        \caption{During training, the loss of different decision models changes with the number of update steps obtained by averaging over three random seeds.}
        \label{fig:Method_Loss}
    \end{figure}

    As depicted in Fig.~\ref{fig:Method_Loss}, which illustrates the loss reduction during training, the loss values of the DT are generally higher towards the end of training compared to the results of the DRWKV model. The overall normalized scores are presented in Table~\ref{tab:Result_D4RL}, where it is observed that the experimental results of the DT are slightly lower than those of the DRWKV model.

    \begin{table}[htbp]
        \centering
        \caption{The results for D4RL datasets, averaged across three random seeds}
        \label{tab:Result_D4RL}
        \begin{tabular}{ccccc}
            \toprule
                Dataset & DT & DRWKV4 & DRWKV5 & DRWKV6 \\
            \midrule
                HalfCheetah-m & 42.53$\pm$0.79 & 42.32$\pm$1.18 &  42.52$\pm$0.74 & 42.59$\pm$0.98 \\
                Hopper-m  & 62.17$\pm$12.24 & 82.05$\pm$13.52 &  69.86$\pm$10.95 &  68.23$\pm$13.23 \\
                Walker-m  & 70.26$\pm$16.31 & 66.94$\pm$13.73 & 68.96$\pm$14.82 &  71.79$\pm$15.86 \\
            \midrule
                HalfCheetah-m-r & 37.80$\pm$7.28 & 40.63$\pm$0.86 & 39.66$\pm$3.15 & 40.16$\pm$1.49 \\
                Hopper-m-r  & 45.76$\pm$20.95 & 66.24$\pm$23.13 & 50.54$\pm$21.02 & 60.55$\pm$24.94 \\
                Walker-m-r & 52.57$\pm$27.22 & 48.34$\pm$25.17 & 63.68$\pm$20.06 & 45.87$\pm$29.44 \\
            \midrule
                HalfCheetah-m-e & 71.14$\pm$25.69 & 84.58$\pm$9.53 & 62.32$\pm$22.40 & 62.93$\pm$23.55 \\
                Hopper-m-e  & 43.90$\pm$23.42 & 52.62$\pm$35.77 & 70.24$\pm$28.12 & 90.36$\pm$28.87 \\
                Walker-m-e & 99.07$\pm$11.27 & 107.55$\pm$0.57 & 107.32$\pm$1.03 & 105.38$\pm$8.63 \\
            \bottomrule
        \end{tabular}
    \end{table}

\begin{figure}[htbp]
        \centering
        \includegraphics[width=0.7\linewidth]{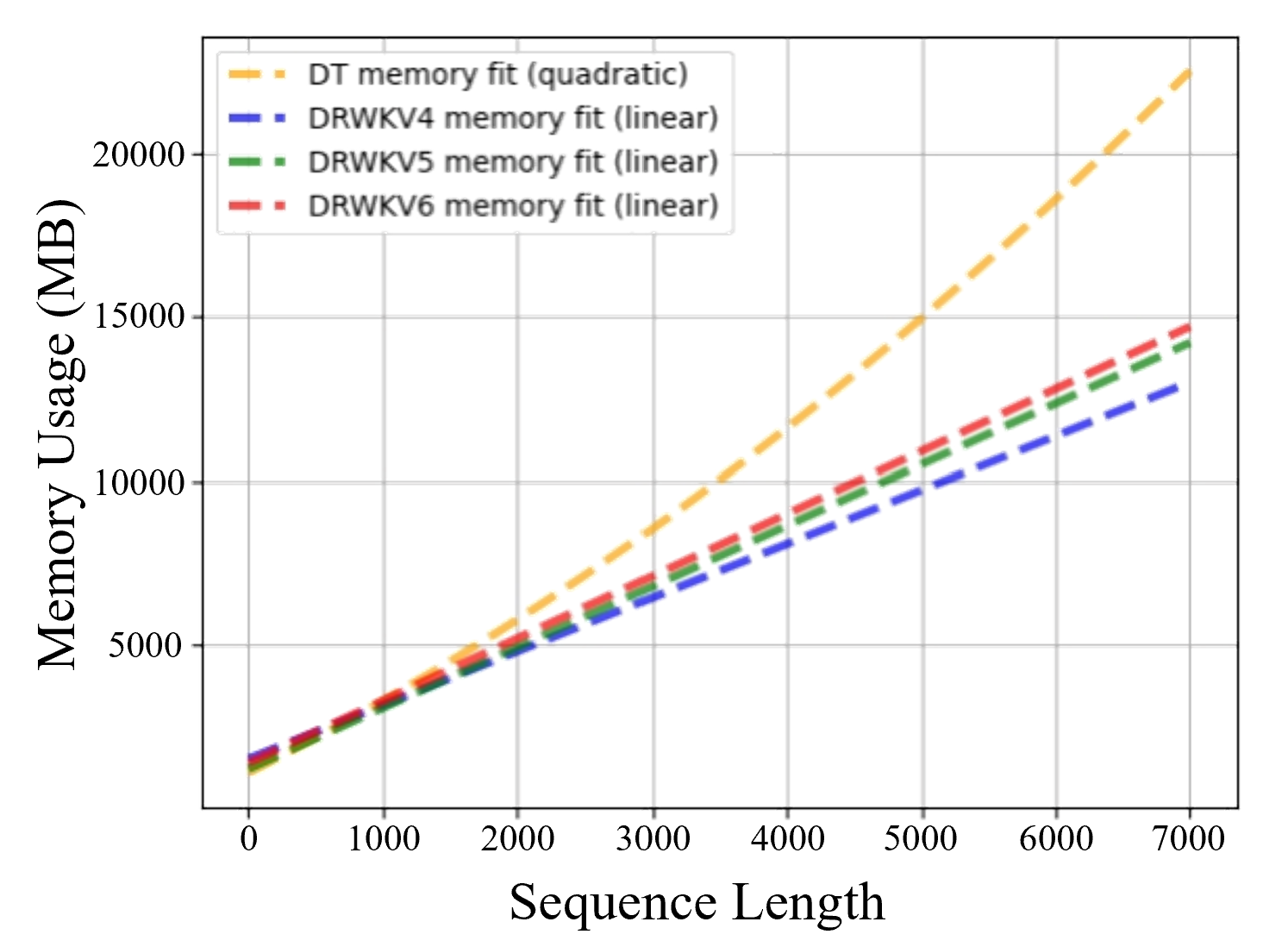}
        \caption{The memory consumption consumption of different decision models varies with the length of the input sequence.}
        \label{fig:Result_MemoryCost}
    \end{figure}

    On the other hand, considering the envisioned application of the model in low-latency scenarios such as robotics, we have documented and analyzed the memory usage during model inference and the computational time required for inference. This consideration is critical as it may limit the deployment of the model in edge computing scenarios where more expensive computational resources are less desirable. Fig.~\ref{fig:Result_MemoryCost} shows the impact of increasing input sequence length on memory usage. 
    
    The sequence length can affect the amount of input data, including time-series data, especially when incorporating multimodal data such as language, images, and audio, which can contain significant information. In applications, robots often require preprocessing of data from multiple sensors and staged multi-model processing, and reducing memory usage is beneficial for enhancing the scalability of robot learning models.

    \begin{figure}[htbp]
        \centering
        \includegraphics[width=0.7\linewidth]{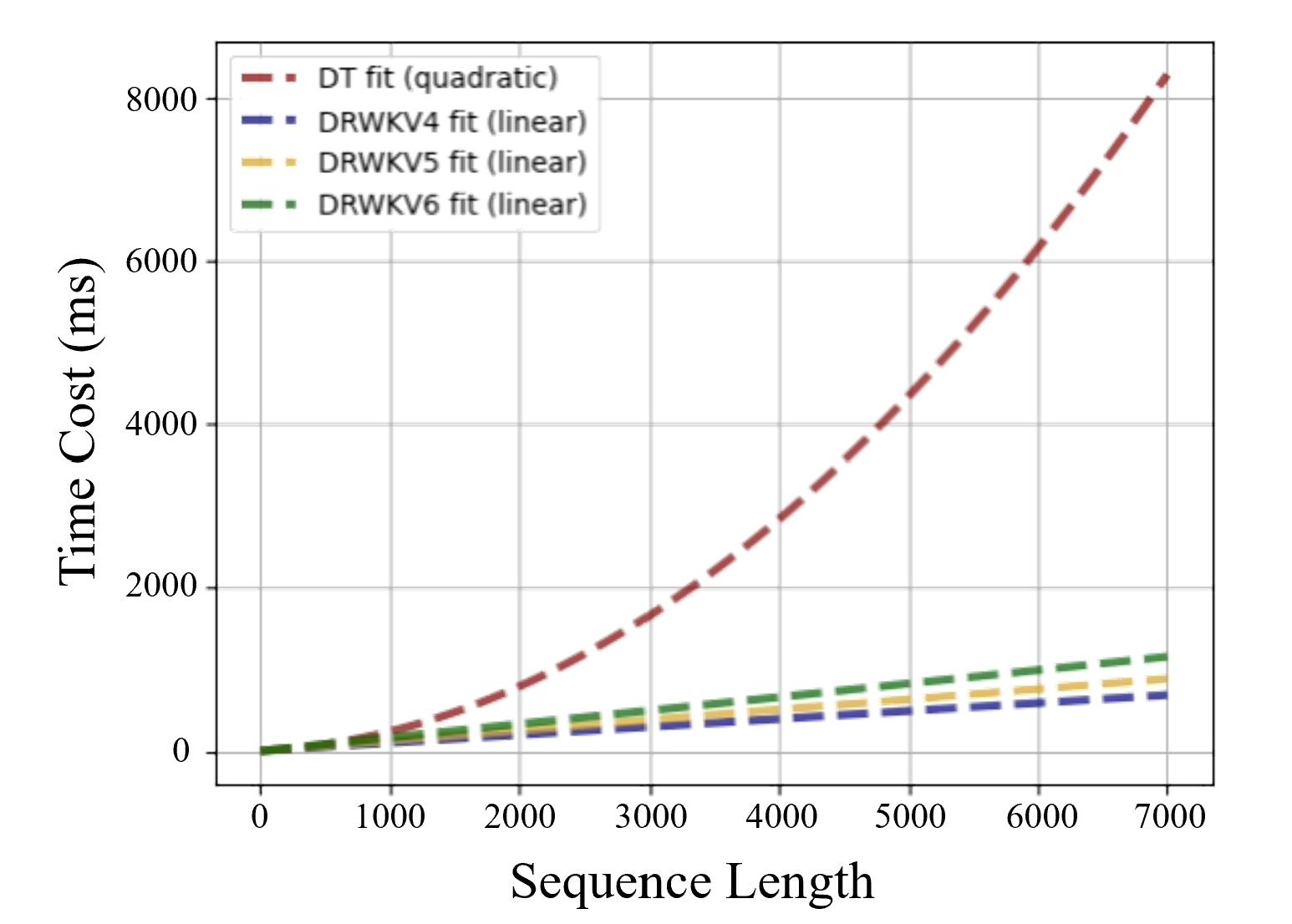}
        \caption{Different decision models' inference time cost varies with the length of the input sequence.}
        \label{fig:Method_TimeCost}
    \end{figure}

    We also analyzed the computational time required for inference, another critical factor for low-latency applications. Fig.~\ref{fig:Method_TimeCost} shows how inference latency varies with input sequence length. As illustrated in Fig.~\ref{fig:Method_TimeCost}, the inference time of the DT model exhibits a quadratic growth with sequence length, which is disadvantageous in scenarios such as autonomous driving and industrial automation where robots need to quickly respond to environmental changes and make decisions, as well as in applications involving human-robot interaction like service robots, where user experience is crucial.

    Therefore, due to its memory efficiency and low-latency inference advantages, the DRWKV model, which possesses linear time and space complexity, can play a significant potential role in real-world robotic applications.

\subsection{Lifelong Robot Learning Result}

    The ROBEL D'Claw~\cite{ahn2020robel} platform operates within the MuJoCo physical simulation environment and has a robotic limb featuring three articulated fingers. Each finger is capable of omnidirectional movement along three distinct axes. The primary experimental task assigned to this robotic system involves precisely rotating a valve to a specified angle, not exceeding 180 degrees. Performance metrics for this task include the collection of joint angles, joint velocities, and the resultant angle of the valve itself. The robot's action space is defined by the range of motion available at its joint articulations.

    To ensure safe and effective manipulation, the reward function has been meticulously designed to consider both the robot's posture and the spatial orientation of the valve. It is important to note that the absolute value of the reward is not inherently significant; instead, it serves as a comparative measure of performance quality, guiding the optimization of manipulation strategies.

    Variability in the experimental setup is introduced solely through valve geometry changes. Despite the superficial similarity across tasks, strategies optimized for a specific task often underperform when applied to others, underscoring the nuanced complexity of manipulation challenges.

    The task design includes a reward structure that provides frequent positive feedback, accelerating the robotic system's learning curve. In subsequent lifelong robot learning experiments, the manipulation tasks were expanded to include twisting ten different shaped valves, as demonstrated in the platform described in~\cite{yang2022evaluations} and illustrated in Fig.~\ref{fig:Intro_Dataset}.

    We use the Soft Actor-Critic (SAC)~\cite{haarnoja2018soft} algorithm for training specialized agents on each task. These agents achieved expert-level performance and were used to generate a comprehensive dataset of expert demonstrations. This dataset is a valuable resource for subsequent experiments in lifelong robot learning, utilizing it as an offline dataset.

\subsubsection{Utilizing Experience Replay in Sequential Learning Across Multi-Tasks} 

    We further delve into applying the proposed model for sequential learning across a series of ten distinct tasks, leveraging an offline dataset for each task. Central to our approach is implementing an Experience Replay Buffer, a memory mechanism designed to store data from previously encountered tasks. This buffer is pivotal in our sequential learning framework, allowing the model to retain and revisit past experiences and facilitating knowledge transfer across tasks.

    The Experience Replay Buffer is maintained with a fixed capacity of 10,000 data entries to ensure a manageable computational footprint. As the model progresses through the sequence of tasks, the buffer is dynamically updated with new experiences. However, it is crucial to note that the total number of stored experiences does not exceed the predefined limit. To achieve this, we employ a strategy that removes older data to make room for newer entries, thereby maintaining a balance between historical and recent knowledge.

    When the model is introduced to a new task, it does not learn in isolation. Instead, the learning process is enriched by combining the new task's data with the diverse experiences stored in the Replay Buffer. This amalgamation of new and past experiences is used to train the model, ensuring that it not only acquires the specifics of the current task but also reinforces and refines the knowledge gained from previous tasks. Doing so encourages the model to develop a more robust and generalizable understanding that can be applied to future tasks.

    The training process with the Replay Buffer is iterative and involves sampling a batch of experiences representative of both the new task and the accumulated knowledge. This sample is then used to update the model's parameters and minimize a loss function reflecting the performance across all tasks encountered. The balance between the new task data and the Replay Buffer content is carefully managed to avoid catastrophic forgetting while promoting positive forward transfer of learning.

    In summary, utilizing the Experience Replay Buffer in our sequential multi-task learning framework is instrumental in achieving a continuous learning paradigm. It allows the model to build upon the knowledge acquired from each task without being constrained by the limitations of task-specific learning. This methodology enhances the model's ability to learn sequentially and ensures that the knowledge is preserved and utilized throughout the learning trajectory.

\subsubsection{Results and Analysis of Lifelong Robot Learning}
    
\begin{figure}[htbp]
        \centering
        \includegraphics[width=0.7\linewidth]{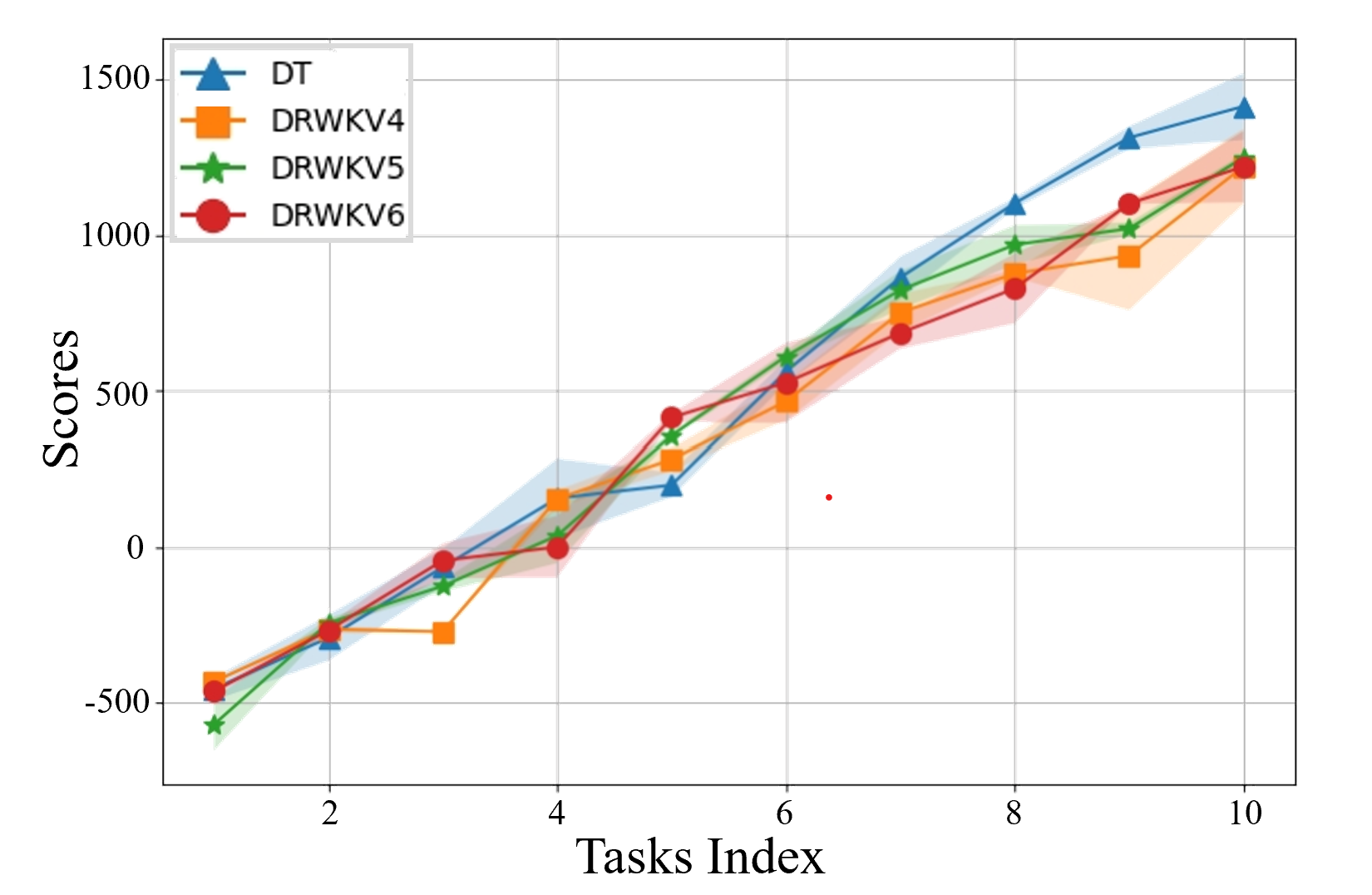}
        \caption{Results of lifelong robotic learning in the D'Claw environment, with scores being the average across all tasks as training progresses according to Eq.~\ref{equ:lifelongRobotLearning}, averaged over data from three random seeds.}
        \label{fig:Result_ERScores}
    \end{figure}

    In this section, we present the results and analysis of our experiments to test the lifelong learning capabilities of robots using the model described in this article. The experiments were conducted across ten tasks associated with its offline dataset to evaluate the model's performance in a lifelong learning context. The outcomes of these experiments are visually represented in Fig.~\ref{fig:Result_ERScores}, which serves as a critical piece of evidence for understanding the model's efficacy in navigating the complexities of sequential task learning.
    
    The primary metric used to gauge the lifelong learning performance of the robot is the average score obtained after training on each task, tested across all ten tasks. This approach allows us to assess not only the immediate learning outcomes of the robot on a given task but also its ability to retain and apply the knowledge gained from previous tasks to new, unseen challenges. The experiments were conducted in diverse settings, including DT, DRWKV4, DRWKV5, and DRWKV6, to comprehensively evaluate the model's capabilities across different environments and task complexities.

    The data analysis in Figure reveals several key insights into the robot's lifelong learning performance. Firstly, the model demonstrates a remarkable ability to adapt to new tasks, as evidenced by the progressive improvement in the average scores with each subsequent task. This trend indicates that the robot is learning the specifics of the current task and effectively leveraging the knowledge acquired from previous tasks to enhance its performance.

    Moreover, the results highlight the model's resilience against catastrophic forgetting, a common challenge in sequential learning scenarios where acquiring new knowledge leads to the erosion of previously learned information. The sustained performance across all ten tasks suggests that the model can balance learning new tasks and preserving the integrity of past knowledge.

    The experimental settings, including DT, DRWKV4, DRWKV5, and DRWKV6, were deliberately chosen to present the robot with various learning challenges. The consistent performance across these diverse settings further underscores the model's versatility and ability to generalize learned concepts across different domains.

    In conclusion, this section's data results and analysis provide compelling evidence of the model's potential in facilitating lifelong robot learning. The ability to learn sequentially across multiple tasks while retaining and applying previous knowledge represents a significant step forward in developing autonomous robots capable of adapting to an ever-changing environment. The insights gained from these experiments not only validate the effectiveness of the proposed model but also open up new avenues for future research in the field of lifelong learning for robots.

\section{Discussions and Conclusion}

    In this work, we examined the capabilities of the recently proposed RWKV for sequence modeling in RL. We acknowledge the decision model's computational intensity and exponentially growing inference time, which pose significant challenges for deployment on edge devices, particularly in robotics. To address these concerns, we pivoted towards the RWKV model, known for its linear complexity, and developed the Decision RWKV. This innovation significantly reduces inference time and memory footprint compared to the DT, making it a more viable option for real-world applications, especially in robotics.

    Building upon this foundation, we integrated the concept of experience replay with the Decision model to devise a lifelong learning algorithm tailored for robots. This algorithm was trained and tested across ten similar offline task datasets, demonstrating commendable performance. The experimental results underscore the potential of the linear-complexity-based DRWKV model in facilitating the learning process for robots and offering insights for future lifelong robot learning endeavors.

    As the development of Large Language Models (LLMs) progresses, the current research trend is moving towards more efficient models. Architectures validated in LLMs, such as RWKV, mamba, and retnet, have shown promise in enhancing models capable of long-sequence memory. These advancements are particularly pivotal in the real-world deployment of models, especially in robotics, where such capabilities can significantly propel robotic operations forward.

    In conclusion, our exploration and findings highlight the significance of pursuing efficient model architectures like DRWKV for lifelong robot learning. The successful integration of experience replays with Decision RWKV marks a step forward in developing algorithms suitable for continuous learning in robots. It aligns with the broader research trend toward efficiency. As we continue to advance in this direction, the prospects for deploying sophisticated models in robotics and other real-world applications appear increasingly promising, paving the way for more adaptive and intelligent systems.

\section*{Acknowledgment} 

    This work was partly supported by the SUSTech Virtual Teaching Lab for Machine Intelligence Design and Learning [Y01331838] and the Science, Technology, and Innovation Commission of Shenzhen Municipality [JCYJ20220818100417038].

\bibliographystyle{unsrt}
\bibliography{References}  
\end{document}